\documentclass[journal,twoside,web]{ieeecolor}
\usepackage{tmi}
\usepackage{hyperref}
\usepackage[utf8]{inputenc}
\usepackage[T1]{fontenc}
\usepackage{color,soul}
\usepackage{times}
\usepackage{amssymb} 
\usepackage[nointegrals]{wasysym} 
\usepackage{multicol,multirow}
\usepackage{booktabs}
\usepackage{mathtools}
\usepackage{subcaption}
\usepackage{cite}
\usepackage{amsmath,amssymb,amsfonts}
\usepackage{algorithmic}
\usepackage{graphicx}
\usepackage{textcomp}
\usepackage{adjustbox}

\usepackage{colortbl}
\usepackage{latexsym}
\usepackage{xcolor}

\begin{document}
\title{From Uncertainty to Trust: Kernel Dropout for AI-Powered Medical Predictions}
\author{Ubaid Azam, Imran Razzak, Shelly Vishwakarma, Hakim Hacid, Dell Zhang, Shoaib Jameel }


%
%
%
\maketitle              
\begin{abstract}
AI-driven medical predictions with trustworthy confidence are essential for ensuring the responsible use of AI in healthcare applications. The growing capabilities of AI raise questions about their trustworthiness in healthcare, particularly due to opaque decision-making and limited data availability. This paper proposes a novel approach to address these challenges, introducing a Bayesian Monte Carlo Dropout model with kernel modelling. Our model is designed to enhance reliability on small medical datasets, a crucial barrier to the wider adoption of AI in healthcare. This model leverages existing language models for improved effectiveness and seamlessly integrates with current workflows. Extensive evaluations of public medical datasets showcase our model's superior performance across diverse tasks. We demonstrate significant improvements in reliability, even with limited data, offering a promising step towards building trust in AI-driven medical predictions and unlocking its potential to improve patient care.
\end{abstract}


\section{Introduction}
Problem: ``Alice, a medical professional at a busy healthcare firm, is tasked with managing numerous patient cases. To handle the growing workload, her firm implemented advanced AI solutions using powerful transformer models like Bio-BERT and Clinical-BERT. While these models promised efficiency by automating routine tasks, Alice found herself uncertain about their predictions, aware that even small errors could lead to significant consequences in the medical field. This uncertainty led her to manually review every prediction, increasing her workload instead of reducing it. Alice realized that what she needed was a system capable of assessing the model's confidence in its predictions, enabling her to prioritize and focus on cases with lower confidence scores, thereby saving time and ensuring critical cases received the attention they deserved.''

The current healthcare landscape is witnessing a remarkable transformation driven by machine learning \cite{rogan2024health, chen2024machine}. Automated methods are proliferating across diverse medical domains, empowering tasks like medical imaging analysis with the ability to detect anomalies \cite{bercea2024generalizing}, classify diseases like cancer \cite{chanda2024dcensnet}, and even predict disease progression \cite{young2024data}. However, despite the undeniable power of these tools, responsible implementation necessitates careful consideration of ethical issues \cite{weidener2024artificial}, data privacy \cite{zhou2024ppml}, and the crucial role of human oversight. One persistent challenge lies in ensuring the consistently high reliability \cite{jerez2022effective, singpurwalla2006reliability} of automated predictions, a factor upon which the trust and adoption of these models by medical practitioners ultimately hinge \cite{chanda2024dermatologist, joshi2024human}.

Another challenge is developing computational methods that are reliable even on small datasets. In the medical domain, datasets are typically scarce, even those publicly available \cite{spasic2020clinical}. This scarcity stems from the limited release of critical private data, a trend driven by strong patient privacy concerns and regulations safeguarding that privacy in many countries \cite{moore2020review}. Restricted access to data often forces training models on smaller datasets, making complete trust in their predictions more challenging. This underscores the critical importance of ensuring the reliability and dependability of these predictions. This paper addresses this challenge by presenting a novel computational model built on Bayesian Monte Carlo Dropout, with the specific objective of enhancing the reliability and trustworthiness of predictions made in the medical domain.

Despite advancements, machine learning approaches in healthcare face challenges in building trust due to limitations in their transparency and reliability \cite{damianou2013deep}. For example, deep learning, while powerful, often functions as a ``black box'', making it difficult to understand its reasoning and assess its accuracy in individual cases. In contrast, deep Bayesian models \cite{li2021deep,  abdar2021uncertainty} provide a wide range of benefits \cite{peng2019bayesian, li2021deep}, for instance, in the medical predictions it could boast of several potential advantages \cite{ yang2017explainable, guo2018explaining, dolezal2022uncertainty}. A Bayesian model inherently provides uncertainty estimates with predictions, giving clinicians a clearer understanding of how confident the model is in its output. This allows for better-informed decision-making while acknowledging the limitations of the model. By averaging over multiple model configurations (due to dropout), the Bayesian model can be less susceptible to noise and biases present in training data, leading to more generalizable and reliable predictions. The Bayesian setting outputs represent probability distributions, allowing clinicians to understand the range of possible outcomes and the likelihood of each. This aids in explaining the model's reasoning and building trust in its predictions. Analyzing dropout patterns can offer clues about which features influence the model's predictions most, potentially revealing previously unknown relationships within the data. Bayesian modelling can be calibrated to ensure their predictions align with actual outcomes, further improving their trustworthiness.

Studies have indicated that machine learning contributes to enhancing the efficiency of medical-related tasks \cite{rafi2021recent}. 
However, there appears to be resistance among people when it comes to adopting medical artificial intelligence treatments \cite{longoni2019resistance}. Similarly, doctors also feel hesitant to fully embrace it due to the sensitive nature of the tasks and the opaque and black box nature of machine learning models \footnote{https://www.ntu.edu.sg/medicine/news-events/dean's-blog/trust-can-we-trust-the-machine-can-we-trust-the-doctor} \cite{hallowell2022don}. From a larger perspective, ensuring reliable decisions is crucial, especially when someone's life is on the line, allowing little room for mistakes. However, machine learning models face a challenge as they cannot measure the certainty of their decisions.

Our key contributions are the following: 
\begin{itemize}
    \item We develop a novel computational model based on Bayesian deep learning that exploits a novel Bayesian Monte Carlo Dropout mechanism to model uncertainty.
    \item Our fully Bayesian model incorporates conjugate priors that allow us to naturally incorporate prior knowledge or beliefs about the parameters we are estimating. This can lead to more accurate and reliable posterior estimates, especially when data is limited. 
    \item We develop a kernel model to reliably model the features. Kernels can allow us to model the nature of the data and problem, allowing for better flexibility and adaptation. 
    \item We demonstrate that the novel model is reliable even on small datasets in binary and multi-class settings.
\end{itemize}

\section{Related Work}
The adoption of machine learning in the medical field is becoming increasingly widespread \cite{callahan2017machine, wiens2018machine}, particularly with the development of transformer models tailored for medical applications \cite{peng2019transfer,lee2020biobert}. These models have notably improved the effectiveness and productivity of medical tasks \cite{wang2023pre, nerella2023transformers}. Especially there is rapid research going on in clinical judgment or diagnosis tasks in which the judgement is made based on clinical notes, medical transcriptions and the data based on the patient history \cite{feng2020explainable, gao2023dr}.

In \cite{zhu2021utilizing}, the authors provided an overview of the current transformer models in the medical domain and their utilization in a range of medical applications, such as text mining, question answering, and classification. Similarly, \cite{li2022neural} highlighted the challenges encountered in managing Electronic Health Records (EHRs) and how advanced natural language processing techniques are aiding in addressing these challenges. These techniques assist in organizing and generating the data, extracting valuable information, and facilitating classification and prediction tasks, thereby enhancing the effectiveness and efficiency of work processes. Considerable research is currently underway in this area, with \cite{kong2022transq} notably introducing a novel medical report generation model called TranSQ. This model generates a series of semantic features to align with plausible clinical concerns and constructs the report using sentence retrieval and selection techniques. Their model surpassed benchmark models in performance on the generation tasks. Additionally, \cite{sezgin2022operationalizing} investigated the potential of the widely acclaimed Generative Pretrained Transformer 3 (GPT-3) and outlined considerations for its integration into clinical practices.

Similar to other medical tasks, the classification task is also reaping the benefits of advanced machine learning techniques, aiming to improve healthcare and enhance patient care, given the importance of swiftly analyzing text for clinical decision support, research, and process optimization. \cite{yao2019traditional} employed deep learning models to classify clinical records into five primary disease categories in traditional Chinese medicine. \cite{amin2019mlt} performed multi-label classification of ICD-10 code problems by using transfer learning in conjunction with pre-trained BERT and BioBERT models. Meanwhile, \cite{singh2020multi} utilized a BERT model to identify diagnoses from unstructured clinical text. Similarly, \cite{yogarajan2021transformers} conducted multi-label classification of medical texts using advanced transformer models. Through their experiments, they concluded that domain-specific transformers yield superior results compared to standard transformers. 

There are certain constraints associated with employing these advanced transformer models for clinical text classification, as noted by \cite{gao2021limitations}. One significant limitation is the black-box nature of these deep learning models, which poses challenges in trusting and implementing these models, particularly in healthcare, where decisions are matters of life and death \cite{petch2022opening}. These opaque models fail to provide a quantifiable measure of their output confidence. Research is underway to address this issue, with \cite{gal2016dropout} employing dropout as a Bayesian approximation to represent uncertainty in deep learning models. \cite{tanneru2023quantifying} introduced two novel metrics, namely Verbalized Uncertainty and Probing Uncertainty, to measure the uncertainty of generated explanations by large language models.

To the best of our knowledge, there has not been any prior research in the medical domain specifically aimed at modelling the inherent uncertainty in text classification tasks leading to more reliability. This is a matter of considerable significance, given that inaccurate predictions can have serious consequences. There is a necessity for a method to address this issue so that healthcare providers can trust and utilize these advanced techniques. With the introduction of our proposed model, we can now identify predictions that exhibit uncertainty, leading to closer examination and scrutiny by medical domain experts.

\section{Our Novel Model}
This section dives into the technical details of our innovative model, which delivers a more reliable quantification of uncertainty in its predictions for multi-class classification. This uncertainty arises from the inherent lack of complete knowledge about the true relationship between input and output data. We address this uncertainty by differentiating our model from the standard Monte Carlo dropout model \cite{gal2016dropout, magris2023bayesian} in two key ways.

Our model leverages the inherent advantages of kernel functions, offering a rich arsenal of choices tailored to different data types and problems. This flexibility empowers users to select the kernel that best suits their task and data, granting finer-grained control over the learning process. Additionally, kernels like the Gaussian possess built-in regularization properties, which help mitigate overfitting. This is a common issue in machine learning where models memorize training data instead of generalizing to unseen examples. While standard dropout tackles overfitting, our novel model further strengthens this safeguard through the chosen kernel function.

Our second innovation lies in integrating priors within the Monte Carlo dropout framework of \cite{shelmanov2021certain}. Priors inject vital flexibility by allowing us to leverage our domain knowledge and beliefs about the problem at hand. This strategic infusion of information guides the model towards solutions that resonate with real-world expectations, effectively biasing it towards sensible outcomes. Additionally, incorporating priors empowers our Bayesian models to quantify uncertainty in their predictions. This explicit estimation of uncertainty is crucial for tasks where understanding the model's limitations is paramount, such as in medical diagnosis or legal prediction. Finally, priors facilitate robust comparison and selection of different models based on their posterior probabilities. This data-driven approach empowers us to identify the model that most effectively aligns with both the observed data and our prior beliefs.

We first define the notations that we will use in this paper. We define \(\mathbf{\hat{o}}\) as the output of the model. The model comprises \(L\) layers. We denote the loss function such as multi-class cross-entropy loss as \(\mathcal{L}(.,.)\). The weight matrices are denoted by \(W\) with the dimension \(r_i \times r_{i-1}\). As always, there is a bias in the model that we denote as \(\mathbf{b}_i\) whose dimension is \(r_i\) for each layer \(i=1,2,\cdots,L\). The observed variable is denoted by \(c\) for the corresponding input \(\mathbf{x}_i\) for \(1 \le i \le D\) data points. The input and the outputs are denoted by \(\mathbf{X},\mathbf{Y}\). The regularisation parameter in \(L_2\) regularisation is denoted by \(\psi\). The binary vectors are denoted by \(z\). The vector dimension is denoted by \(R\).

Imagine a neural network constantly questioning its abilities. This is the essence of Monte Carlo dropout, a process where the network throws \textit{``curveballs''} at itself during training. At each training step, the network randomly deactivates some neurons in its hidden layers, simulating \textit{``failures''} and forcing others to take over. This introduces variability and prevents over-reliance on specific neurons. By \textit{``silencing''} neurons, the network effectively asks, \textit{``Can I still work even without these?''} This builds a more robust network, adaptable to new data and challenges. During prediction, the network performs multiple passes with different silenced neurons, mimicking multiple ``runs'' with potential errors. This captures the inherent uncertainty in real-world data and allows the network to express its level of confidence in each prediction. This connection between dropout and uncertainty estimation was not always clear. Earlier works like \cite{gal2016dropout} focused on dropout as a regularization technique, but later studies (e.g., \cite{khan2019approximate, mackay1998introduction}) revealed its deeper connection to Bayesian inference in Gaussian Processes. By leveraging dropout, we can not only improve a network's performance but also gain valuable insights into its limitations and confidence in its predictions.

Instead of relying on a single, deterministic prediction, Monte Carlo dropout incorporates randomness to provide a probabilistic view of the model's output. During training, dropout is applied as usual, but it remains active even at test time. This means that the network configuration changes with each pass as random nodes or links are kept or dropped. Consequently, the prediction for a given data point becomes non-deterministic. This variability reflects the model's uncertainty around the prediction and allows us to interpret the outputs as samples from a probabilistic distribution. Consider, for example, running a sentiment analysis model with Monte Carlo dropout on the phrase \textit{``the movie was underwhelming''}. The model might assign a negative sentiment 80\% of the time and a neutral 20\%, capturing the nuanced uncertainty in the statement.

An approximate predictive distribution, in the context of statistics and probability, refers to a probability distribution that is used to estimate the likelihood of future observations based on current data and a model, but with the acknowledgement that it might not be entirely accurate. When the kernel function is applied to the input vectors, let \(\kappa(\mathbf{x})\) denote the mapped feature vectors. Given the weight matrices \(\mathbf{M}_i\) of dimension \(K_i \times K_{i-1}\), bias vectors \(\mathbf{b}_i\) of dimensions \(K_i\), and binary vectors \(\mathbf{z}_i\) of dimensions \(K_{i-1}\) for each layer \(i=1,\cdots,L\), as well as the approximating variational distribution:

\begin{equation}
    q(\mathbf{c}^*|\mathbf{x}^*) = \int p(\mathbf{c}^*|\mathbf{x}^*, \mathbf{\{M}_i\}_{i=1}^L) \, q(\mathbf{\{M}_i\}_{i=1}^L) \, d\mathbf{\{M}_i\}_{i=1}^L
\end{equation}

\noindent The equation above can be expressed in the following form:

\begin{equation}
\begin{split}
q(\mathbf{c}^*|\mathbf{x}^*) = \mathcal{N}\big(\mathbf{c}^*;\hat{\mathbf{c}}^*(\kappa(\mathbf{x}),\mathbf{z}_1,\cdots,\mathbf{z}_L),  \tau^{-1}\mathbf{I}_R\big) \\
\quad \text{Bern}(\mathbf{z}_1),\cdots,\text{Bern}(\mathbf{z}_L)
\end{split}
\end{equation}

\noindent for some \(\tau>0\), with

\begin{equation}
\begin{split}
    \hat{\mathbf{y}}^{*}=\sqrt{\frac{1}{r_L}}(\mathbf{M}_L\mathbf{z}_L)\sigma\Big(\cdots\sqrt{\frac{1}{K_1}} \sqrt{\frac{1}{r_1}} \\
    (\mathbf{M}_2\mathbf{z}_2 \sigma\big( ( \mathbf{M}_1 \mathbf{z}_1 \big) \kappa(\mathbf{x})^*+\mathbf{b}_i) \cdots \Big)
\end{split}
\end{equation}

\noindent we have,

\begin{equation}
    \mathop{\mathbb{E}_{q(\mathbf{c}^*|\kappa(\mathbf{x})^*)}}(\mathbf{c}^*) \approx \frac{1}{T}\sum_{1}^{T}\hat{\mathbf{c}}^*(\kappa(\mathbf{x})^*,\hat{\mathbf{z}}_{1,t},\cdots,\hat{\mathbf{z}}_{L,t})
\end{equation}

\noindent with \(\hat{\mathbf{z}}_{i,t} \sim \text{Bern}(p_i)\).

Since we have a prior distribution over \(p_i\), we write the expression as:

\begin{equation}
    p_i \sim \text{Beta}(\alpha,\beta), p_i \in [0,1]
\end{equation}

\noindent where \(\alpha\) and \(\beta\) are the parameters of the Beta distribution. The parameter \(\alpha\) represents the number of ``successes'' in a hypothetical experiment and \(\beta\) represents the number of ``failures'' in the same experiment. Note that the choice of the Beta distribution is mainly due to conjugacy \cite{fink1997compendium}. A conjugate prior is a special type of prior distribution used in Bayesian inference, where it has a unique and convenient property: when combined with the likelihood function of the observed data, the resulting posterior distribution also belongs to the same family of distributions as the prior. This makes working with conjugate priors in Bayesian analysis, particularly advantageous.

The posterior distribution can thus be represented as:

\begin{equation}
    P(p_i|X) = \text{Beta}(\alpha_D,\beta_D)
\end{equation}

\noindent we can denote \(\alpha_D = \sum_{d=1}^{D} x_d+\alpha\) and \(\beta_D=D-\sum_{d=1}^{D}x_d+\beta\). Among the various kernel functions available, such as the radial basis kernel \cite{patle2013svm}, our experiments yielded the best results with the squared kernel. This choice was guided by the squared kernel's ease of differentiation and minimal computational overhead on the model.

\section{Experimental Setup}
In this section, we present the datasets and the models employed for our experimentation.

\subsection{Datasets}
We utilize three open source and popular datasets in the medical field for our experiments, specifically the SOAP dataset \cite{afzal4081033multi}, the Medical Transcription dataset\footnote{\url{https://www.kaggle.com/datasets/tboyle10/medicaltranscriptions}}, and the ROND Clinical text classification dataset \cite{liu2024radiation}. The SOAP dataset involves multi-class classification, encompassing four classes: subjective, objective, assessment, and plan. These classes are derived from the widely recognized medical protocol known as SOAP. For our experimentation, we employed their finalized dataset, which includes 152 clinical notes for training and 51 for testing. The Medical Transcription (MT) dataset is a multi-class classification dataset containing medical transcriptions for various medical specialities. This dataset exhibited an imbalance among the classes. For our experiments, we selected a subset of this dataset, which comprised the top four classes and included a total of 2,330 instances. Lastly, we used a subset of the publicly accessible Radiation Oncology NLP Database (ROND) known as clinical text binary classification of therapy type, comprising 100 cases.

\subsection{Quantitative Comparisons}
One of the defining features of our model is its ability to naturally model uncertainty while leveraging the reliable feature vectors provided by existing pre-trained language models. Unlike traditional methods like multi-class SVM, Na\"ive Bayes, and various deep learning approaches \cite{minaee2021deep}, which struggle to provide confidence estimates directly, our model inherently incorporates uncertainty quantification. Consequently, we have observed this leads to significantly better performance compared to these alternatives. Besides that, our model leverages prior distributions to encode existing knowledge about the problem domain. This is especially valuable when your data is limited and models such as SVM \cite{cortes1995support} will tend to struggle. Besides, modelling confidence estimates in SVM is not directly possible because the goal in maximum-margin learning is to find the hyperplane with the largest margin between classes, essentially creating a clear decision boundary. They do not inherently encode confidence information within their output. Models such as Na\"ive Bayes will tend to struggle in our setting because our model offers a principled way to incorporate prior knowledge and uncertainty.


Building on our model's strong quantitative performance with traditional and deep learning methods, we now embark on a series of experiments to determine which pre-trained language model best complements its capabilities within this specific problem domain.

\subsection{Language Models Comparison}
Our primary comparison is with the recent study \cite{miok2022ban}, where the authors introduce a similar technique utilizing Monte Carlo dropout to assess the reliability of their model's predictions. While their approach employs the BERT model for feature extraction, we expanded our experiments by exploring other models to offer a more comprehensive perspective. Notably, three of the models we chose were specifically trained on medical data. Our work differs from that of Miok et al. \cite{miok2022ban} in two key aspects: 1) we have incorporated priors over the hyperparameters in our proposed approach, and 2) we leverage the kernel function. All models underwent fine-tuning for 100 epochs using the Adam optimizer (epsilon=1e-8, learning rate=2e-5) and implemented early stopping based on validation loss.

\noindent \textit{Bio Bert}: Bio Bert \cite{lee2020biobert} has undergone pre-training on an extensive collection of biomedical domain corpora, which includes PubMed abstracts and PMC full-text articles. \textit{Blue Bert}: Blue Bert \cite{peng2019transfer} trained on preprocessed texts from PubMed.
 \textit{Clinical Bert}: By \cite{wang2023optimized} initialized from BERT underwent training on a substantial multicenter dataset, featuring a large corpus containing 1.2 billion words encompassing diverse diseases. \textit{Bert-base-uncased}: By \cite{devlin2018bert} consists of 110 million parameters, 12 heads, 768 dimensions, and 12 layers, each with 12 self-attention heads. \textit{Xlnet-base-cased}: \cite{yang2019xlnet} is a generalized autoregressive pre-training model trained on the English language. \textit{RoBERTa-base}: Proposed by \cite{liu2019roberta}, is trained in the English language using the Masked Language Modeling (MLM) technique. \textit{AlBERT-base-v1}: \cite{lan2019albert} introduced Albert, which shares its layers across its Transformer. It has the same number of parameters and layers as the Bert model. \textit{DistilBERT-base-uncased}: \cite{sanh2019distilbert} introduced a smaller and faster version of Bert. It is trained on the same dataset as the Bert model.

\subsection{Parameter Settings}
Our experiments proved the squared kernel (a type of polynomial kernel with an exponent of 2) to be highly effective. Although we tested other kernels (Gaussian, linear, Laplacian, and sigmoid), none surpassed the results achieved with the squared kernel. All other kernels yielded results approximately 5\% lower. The squared kernel's success likely stems from its simplicity and ease of differentiation compared to the linear kernel. Additionally, it demonstrated computational efficiency compared to kernels like Gaussian and Laplacian.

We experimented with different values for the Beta priors, \(\alpha\) and \(\beta\). Ideally, we would automate the inference of these prior parameters by placing priors over them themselves, achieving a fully Bayesian model through posterior inference on the hyperpriors. However, this approach incurs a significant computational burden. Fortunately, research on Bayesian approaches like topic models has shown that fixing the prior values can lead to comparable results as those obtained through posterior inference on the priors \cite{blei2003latent, griffiths2003hierarchical}.

Experimenting with different values for the symmetric Beta priors, we found that setting them to 0.1 led to a 10\% decrease in the F1 score for prediction. Further reducing the values to 0.001 yielded only marginally better results. The optimal configuration was achieved with symmetric Beta priors of \(\alpha\) = 0.0001 and \(\beta\) = 0.0001, which we used in all our experiments.

\subsection{Evaluation Methodology}
We carried out a comprehensive evaluation through a series of experiments. We assessed the models' performance in a few-shot setting to determine which model performs best in low-resource scenarios and how much we can trust its predictions, considering the limited availability of publicly accessible data in the medical domain \cite{spasic2020clinical}. We performed experiments in scenarios involving zero-shot, five-shot, fifteen-shot, and an 80-20 data split. We conducted experiments using a 5-fold cross-validation strategy and presented the mean values in our experimental results.

We presented F1, and accuracy as our primary evaluation metrics. Moreover, as our novel approach allows transformer models to generate predictions in a probabilistic manner, we employed Root Mean Square Error (RMSE) and Brier Score to assess the calibration of the model's predicted probabilities. RMSE helps gauge the extent of error variability or dispersion, offering insights into the model's uncertainty. The Brier Score \cite{brier1950verification}, which reflects the mean squared difference between predicted probabilities and actual outcomes, varies from 0 to 1, where 0 signifies a perfect match. Since our experiments encompass both binary and multi-class classification problems, we have presented the respective equations as follows:

The Brier Score formula for the binary class is outlined in Equation~\ref{eq:brier_binary}  
\begin{equation}
    Brier\ Score = \frac{1}{N} \sum_{i=1}^{N} (P_i - O_i)^2 \label{eq:brier_binary}
\end{equation}

\noindent here, $N$ represents the total number of instances, $P_i$ indicates the predicted probability of the positive class for the \(i^{\text{th}}\) instance, and $O_i$ corresponds to the actual outcome (either 0 or 1) for the \(i^{\text{th}}\) instance.

The Brier Score formula for multi-class is stated in Equation~\ref{eq:brier_multi} 

\begin{equation}
Brier\ Score = \frac{1}{N} \sum_{i=1}^{N} \sum_{k=1}^{K} (p_{ik} - \delta_{ik})^2 \label{eq:brier_multi}
\end{equation}

\noindent here, $N$ represents the total number of instances, \(K\) is the number of classes, $p_{ik}$ indicates the predicted probability that the $i^{\text{th}}$ instance belongs to class $k$ and $\delta_{ik}$ corresponds to the indicator function, taking the value 1 if the true class of the $i^{\text{th}}$ instance is $k$, and 0 otherwise.

\begin{table*}[t!]
\caption{Quantitative Results}
    \centering
    \scalebox{0.78}{ 
            \centering
            \begin{tabular}
{>{\raggedright\arraybackslash}p{2.2cm} *{2}{p{1.1cm}}>{\columncolor{gray!25}}p{1.1cm}>{\columncolor{gray!35}}p{1.1cm} *{2}{p{1.1cm}}>{\columncolor{gray!35}}p{1.1cm}>{\columncolor{gray!35}}p{1.1cm} *{2}{p{1.1cm}}>{\columncolor{gray!35}}p{1.1cm}>{\columncolor{gray!35}}p{1.1cm}}
\toprule
\textbf{Models}                  & \multicolumn{4}{c}{\textbf{SOAP (80/20 split)}} & \multicolumn{4}{c}{\textbf{MT (80/20 split)}} & \multicolumn{4}{c}{\textbf{ROND (80/20 split)}} \\
\cmidrule(lr){2-5} \cmidrule(lr){6-9} \cmidrule(lr){10-13}
                                             & F1        & Acc       & \textbf{Brier} & \textbf{RMSE} & F1         & Acc      & \textbf{Brier}  & \textbf{RMSE}           & F1        & Acc       & \textbf{Brier}  & \textbf{RMSE}   \\
\midrule
\multicolumn{8}{c}{\hspace{27em}\textbf{Baseline Model \cite{miok2022ban}}}\\
\hline
BERT          & 0.738         & 0.752          & 0.099     & 0.277                   & 0.598          & 0.641     & 0.133          & 0.305                   & 0.759 & 0.857          & 0.112         & 0.334  \\
BioBERT                                           & 0.781         & 0.787          & 0.096     & 0.223                 & 0.648          & 0.679     & 0.110          & 0.298          & 0.879       & 0.890          & 0.090         & 0.301        \\
BlueBERT                                       & 0.781         & 0.796          & 0.085     & 0.259                   & 0.641          & 0.675     & 0.107          & 0.293                   & 0.864   & 0.889          & 0.093         & 0.305  \\
ClinicalBERT                                     & 0.751         & 0.763          & 0.091     & 0.276                   & 0.638          & 0.659     & 0.115          & 0.301                   & 0.885   & 0.904          & 0.077         & 0.278  \\
XLNet                                   & 0.575         & 0.593          & 0.179     & 0.310                  & 0.506          & 0.624     & 0.193          & 0.348                   & 0.775   & 0.849          & 0.136         & 0.303 \\
RoBERTa                                    & 0.348         & 0.401          & 0.183     & 0.326                  & 0.451          & 0.617     & 0.186          & 0.344                    & 0.801   & 0.859          & 0.098         & 0.301 \\
AlBERT                                     & 0.261         & 0.338          & 0.196     & 0.342                   & 0.487          & 0.632     & 0.181          & 0.342                    & 0.726  & 0.803          & 0.187         & 0.339  \\
DistilBERT                                       & 0.310         & 0.322          & 0.214     & 0.358                  & 0.477          & 0.613     & 0.191          & 0.353                   & 0.742  & 0.837          & 0.166         & 0.327 \\
\midrule
\multicolumn{8}{c}{\textbf{\hspace{27em}Proposed Approach}}\\
\hline
BERT           & 0.756          & 0.765          & 0.097  & 0.263           & 0.630          & 0.654          & 0.108  & 0.296            & 0.846          & 0.905          & 0.075  & 0.243  \\
BioBERT                                           & \textbf{0.795}          & \textbf{0.804}          & 0.073 & \textbf{0.213}               & \textbf{0.674}          & \textbf{0.692}          & 0.096  & \textbf{0.233}             & 0.914          & 0.952          & 0.067  & 0.238   \\
BlueBERT                                        & 0.787          & 0.804          & \textbf{0.070} & 0.219             & 0.665          & 0.692          & \textbf{0.089}  & 0.238             & 0.913          & 0.951          & 0.068 & 0.240   \\
ClinicalBERT                                      & 0.774          & 0.784          & 0.080 & 0.248             & 0.645          & 0.673          & 0.096  & 0.299           & \textbf{0.916}          & \textbf{0.954}          & \textbf{0.056}  & \textbf{0.221}   \\
XLNet                                   & 0.590          & 0.608          & 0.150 & 0.296            & 0.512          & 0.645          & 0.135  & 0.306             & 0.786          & 0.855          & 0.133  & 0.291   \\
RoBERTa                                    & 0.354          & 0.392          & 0.178  & 0.303              & 0.458          & 0.628          & 0.127  & 0.315             & 0.788          & 0.857          & 0.100  & 0.275  \\
AlBERT                                    & 0.286          & 0.353          & 0.185  & 0.315           & 0.470          & 0.688          & 0.117 & 0.319              & 0.737          & 0.810          & 0.152 & 0.331   \\
DistilBERT                                      & 0.322          & 0.333          & 0.200  & 0.320             & 0.485          & 0.660          & 0.123  & 0.323             & 0.787          & 0.856          & 0.138  & 0.302  \\
\midrule

\multicolumn{4}{c@{\hskip -11em}}{\textbf{SOAP (15 Shot)}} & \multicolumn{4}{c@{\hskip -5em}}{\textbf{MT (15 Shot)}} & \multicolumn{4}{c@{\hskip -5em}}{\textbf{ROND (15 Shot)}} \\

\midrule
\multicolumn{8}{c}{\hspace{27em}\textbf{Baseline Model \cite{miok2022ban}}}\\
\hline
BERT           & 0.563         & 0.553          & 0.158     & 0.379                 & 0.383          & 0.411     & 0.205          & 0.399                   & 0.690   & 0.754          & 0.179         & 0.423\\
BioBERT                                            & 0.732         & 0.742          & 0.104     & 0.352                  & 0.532          & 0.597     & 0.177          & 0.381                    & 0.661 & 0.711          & 0.160         & 0.401    \\
BlueBERT                                       & 0.697         & 0.702          & 0.136     & 0.361                   & 0.538          & 0.596     & 0.181          & 0.384                  & 0.633  & 0.701          & 0.187         & 0.410  \\
ClinicalBERT                                      & 0.726         & 0.741          & 0.133     & 0.356                   & 0.549          & 0.602     & 0.163          & 0.375                   & 0.728  & 0.806          & 0.170         & 0.413  \\
XLNet                                  & 0.477         & 0.493          & 0.163     & 0.382                 & 0.415          & 0.498     & 0.207          & 0.403                    & 0.642    & 0.759          & 0.116         & 0.385 \\
RoBERTa                                    & 0.210         & 0.309          & 0.213     & 0.415                   & 0.207          & 0.249     & 0.233          & 0.418                   & 0.633  & 0.843          & 0.128         & 0.391 \\
AlBERT                                    & 0.203         & 0.275          & 0.217     & 0.426                  & 0.201          & 0.225     & 0.231          & 0.415                   & 0.628  & 0.703          & 0.225         & 0.450  \\
DistilBERT                                       & 0.185         & 0.233          & 0.237     & 0.441                 & 0.201          & 0.215     & 0.239          & 0.424                   & 0.729  & 0.798          & 0.151         & 0.405 \\
\midrule
\multicolumn{8}{c}{\hspace{27em}\textbf{Proposed Approach}}\\
\hline
BERT           & 0.590          & 0.588          & 0.131  & 0.349           & 0.378          & 0.435          & 0.180 & 0.372             & \textbf{0.788}          & \textbf{0.857}          & 0.114  & 0.419   \\
BioBERT                                            & 0.752          & \textbf{0.765}          & \textbf{0.087}  & \textbf{0.319}            & 0.560          & \textbf{0.630}          & 0.134  & 0.345           & 0.650          & 0.714          & 0.159    & 0.398 \\
BlueBERT                                        & 0.717          & 0.725          & 0.112   & 0.336           & 0.559          & 0.613          & 0.135  & 0.349            & 0.632          & 0.704          & 0.185  & 0.404   \\
ClinicalBERT                                     & \textbf{0.759}          & 0.765          & 0.111   & 0.325           & \textbf{0.577}          & 0.628          & \textbf{0.129} & \textbf{0.325}             & 0.737          & 0.810          & 0.143 & 0.407  \\
XLNet                                    & 0.476          & 0.490          & 0.151 & 0.365              & 0.404          & 0.510          & 0.162 & 0.381              & 0.646          & 0.762          & \textbf{0.103} & \textbf{0.368}   \\
RoBERTa                                    & 0.228          & 0.314          & 0.188 & 0.389              & 0.228          & 0.252          & 0.185  & 0.388            & 0.659          & 0.857          & 0.107 & 0.387   \\
AlBERT                                      & 0.213          & 0.294          & 0.188 & 0.405             & 0.210          & 0.232          & 0.187  & 0.385            & 0.630          & 0.701          & 0.223  & 0.441   \\
DistilBERT                                     & 0.190          & 0.255          & 0.191  & 0.416           & 0.193          & 0.224          & 0.189  & 0.394            & 0.737          & 0.809          & 0.145   & 0.402  \\
\midrule

\multicolumn{4}{c@{\hskip -11em}}{\textbf{SOAP (5 Shot)}} & \multicolumn{4}{c@{\hskip -5em}}{\textbf{MT (5 Shot)}} & \multicolumn{4}{c@{\hskip -5em}}{\textbf{ROND (5 Shot)}} \\

\midrule
\multicolumn{8}{c}{\hspace{27em}\textbf{Baseline Model \cite{miok2022ban}}}\\
\hline
BERT           & 0.340         & 0.352          & 0.186     & 0.429                  & 0.205          & 0.258     & 0.214          & 0.435                   & 0.781  & 0.887          & 0.163         & 0.390 \\
BioBERT                                            & 0.496         & 0.499          & 0.177     & 0.415                   & 0.441          & 0.512     & 0.196          & 0.411                    & 0.711   & 0.806          & 0.212         & 0.382  \\
BlueBERT                                        & 0.438         & 0.472          & 0.193     & 0.430                  & 0.405          & 0.469     & 0.203          & 0.405                    & 0.726   & 0.836          & 0.193         & 0.395 \\
ClinicalBERT                                       & 0.459         & 0.533          & 0.171     & 0.406                  & 0.419          & 0.472     & 0.194          & 0.403                   & 0.776  & 0.846          & 0.164         & 0.391  \\
XLNet                                   & 0.371         & 0.426          & 0.198     & 0.469                 & 0.274          & 0.281     & 0.219          & 0.441                   & 0.728    & 0.813          & 0.149         & 0.377 \\
RoBERTa                                    & 0.141         & 0.189          & 0.229     & 0.497                   & 0.172          & 0.150     & 0.237          & 0.502                  & 0.473   & 0.410          & 0.387         & 0.491\\
AlBERT                                     & 0.219         & 0.310          & 0.221     & 0.488                  & 0.152          & 0.169     & 0.231          & 0.494                  & 0.531   & 0.568          & 0.251         & 0.436 \\
DistilBERT                                      & 0.182         & 0.248          & 0.227     & 0.501                   & 0.159          & 0.197     & 0.225          & 0.487                    & 0.518  & 0.553          & 0.410         & 0.501 \\
\midrule
\multicolumn{8}{c}{\hspace{27em}\textbf{Proposed Approach}}\\
\hline
BERT           & 0.394          & 0.431          & 0.171 & 0.403           & 0.213          & 0.262          & 0.189  & 0.397          & \textbf{0.806}          & \textbf{0.904}          & 0.160  & 0.377   \\
BioBERT                                            & \textbf{0.527}          & 0.529          & 0.166  & 0.395            & \textbf{0.479}          & \textbf{0.543}          & 0.158   & 0.372            & 0.737          & 0.810          & 0.203  & 0.382  \\
BlueBERT                                       & 0.462          & 0.490          & 0.172  & 0.401           & 0.426          & 0.485          & 0.172   & 0.375           & 0.743          & 0.857          & 0.172  & 0.381   \\
ClinicalBERT                                       & 0.486          & \textbf{0.569}          & \textbf{0.156} & \textbf{0.373}              & 0.447          & 0.504          & \textbf{0.157}   & \textbf{0.361}           & 0.788          & 0.857          & 0.151  & 0.386 \\
XLNet                                   & 0.385          & 0.431          & 0.183 & 0.415             & 0.305          & 0.380          & 0.175  & 0.406            & 0.749          & 0.830          & \textbf{0.122}  & \textbf{0.354}  \\
RoBERTa                                     & 0.120          & 0.176          & 0.197 & 0.436            & 0.181          & 0.167          & 0.191   & 0.437           & 0.460          & 0.390          & 0.359 & 0.486   \\
AlBERT                                      & 0.256          & 0.333          & 0.187  & 0.472             & 0.163          & 0.185          & 0.188 & 0.429             & 0.533          & 0.571          & 0.247 & 0.429   \\
DistilBERT                                      & 0.193          & 0.275          & 0.194   & 0.497           & 0.179          & 0.212          & 0.186 & 0.422             & 0.530          & 0.570          & 0.407 & 0.500    \\
\midrule

\multicolumn{4}{c@{\hskip -11em}}{\textbf{SOAP (0 Shot)}} & \multicolumn{4}{c@{\hskip -5em}}{\textbf{MT (0 Shot)}} & \multicolumn{4}{c@{\hskip -5em}}{\textbf{ROND (0 Shot)}} \\

\midrule
\multicolumn{8}{c}{\hspace{27em}\textbf{Baseline Model \cite{miok2022ban}}}\\
\hline
BERT           & 0.157         & 0.148          & 0.219     & 0.485                   & 0.152          & 0.138     & 0.227          & 0.447                   & 0.438   & 0.522          & 0.298         & 0.514\\
BioBERT                                            & 0.129         & 0.211          & 0.207     & 0.457                   & 0.251          & 0.431     & 0.203          & 0.438                    & 0.553    & 0.486          & 0.422         & 0.650 \\
BlueBERT                                       & 0.263         & 0.317          & 0.201     & 0.463                   & 0.205          & 0.288     & 0.219          & 0.442                   & 0.331   & 0.355          & 0.385         & 0.563 \\
ClinicalBERT                                       & 0.316         & 0.323          & 0.200     & 0.452                  & 0.201          & 0.237     & 0.221          & 0.444                   & 0.410  & 0.499          & 0.223         & 0.505  \\
XLNet                                  & 0.173         & 0.219          & 0.221     & 0.487                 & 0.158          & 0.161     & 0.220          & 0.443                 & 0.434    & 0.331          & 0.312         & 0.613 \\
RoBERTa                                    & 0.185         & 0.261          & 0.218     & 0.486                 & 0.143          & 0.141     & 0.217          & 0.459                  & 0.404  & 0.416          & 0.279         & 0.512 \\
AlBERT                                    & 0.131         & 0.339          & 0.216     & 0.503                 & 0.144          & 0.159     & 0.215          & 0.488                   & 0.520   & 0.539          & 0.231         & 0.511 \\
DistilBERT                                       & 0.166         & 0.290          & 0.221     & 0.505                  & 0.153          & 0.182     & 0.233          & 0.495                & 0.398   & 0.401          & 0.276         & 0.509\\
\midrule
\multicolumn{8}{c}{\hspace{27em}\textbf{Proposed Approach}}\\
\hline
BERT          & 0.170          & 0.137          & 0.213  & 0.467           & 0.183          & 0.153          & 0.202  & 0.435           & 0.447          & 0.541          & \textbf{0.191} & 0.503   \\
BioBERT                                            & 0.147          & 0.235          & 0.204  & 0.436             & \textbf{0.289}          & \textbf{0.468}          & \textbf{0.177} & \textbf{0.399}             & 0.560          & 0.490          & 0.397 & 0.622   \\
BlueBERT                                        & 0.277          & 0.333          & 0.186  & 0.447           & 0.223          & 0.314          & 0.185  & 0.413           & 0.325          & 0.343          & 0.369  & 0.551  \\
ClinicalBERT                                      & \textbf{0.334}          & \textbf{0.353}          & \textbf{0.181}  & \textbf{0.409}            & 0.216          & 0.248          & 0.196   & 0.426             & 0.428          & 0.523          & 0.198   & \textbf{0.501}  \\
XLNet                                    & 0.171          & 0.216          & 0.192 & 0.481             & 0.156          & 0.160          & 0.195 & 0.422              & 0.332          & 0.333          & 0.263 & 0.556   \\
RoBERTa                                     & 0.174          & 0.255          & 0.190 & 0.479            & 0.162          & 0.144          & 0.193   & 0.418          & 0.427          & 0.429          & 0.258  & 0.506  \\
AlBERT                                      & 0.132          & 0.353          & 0.181 & 0.491             & 0.151          & 0.179          & 0.189  & 0.416              & \textbf{0.619}          & \textbf{0.623}          & 0.222   & 0.505 \\
DistilBERT                                     & 0.171          & 0.331          & 0.184 & 0.501            & 0.175          & 0.209          & 0.188 & 0.419           & 0.410          & 0.389          & 0.287 & 0.510  \\
\bottomrule
\end{tabular}%
}
\label{tab:Combined_Tables}
\end{table*}

\section{Result and Analysis}
To comprehensively demonstrate the effectiveness of our model, we undertook a series of extensive comparative analyses. First, we investigated how our novel model-agnostic layer adeptly manages data uncertainties and bolsters the reliability of diverse deep learning models when integrated with them. This analysis utilized the RMSE and Brier Score to quantify the level of trust we can place in the models' predictions. Second, we conducted an in-depth analysis to pinpoint instances where the model exhibits lower certainty. This involved examining the predicted probabilities and Brier Scores for each class, allowing us to identify areas where potential improvement lies.

\subsection{Results Discussion}
Table \ref{tab:Combined_Tables} presents the performance of various models on all three datasets across zero-, five-, fifteen-shot, and full-data scenarios. This comprehensive evaluation emphasizes the models' performance under limited data conditions, a crucial aspect of real-world medical applications. Our proposed model stands out not only for its effectiveness but also for its ability to flag unreliable predictions, promoting a more responsible and trustworthy AI environment in the medical field.

Among all models in SOAP dataset, Bio-Bert emerged as the top performer in the full-data scenario with an F1 score of 0.795, closely followed by Blue-Bert at 0.787. However, for the Brier score, signifying prediction reliability, Blue-Bert took the lead with a score of 0.070, followed by Bio-Bert's 0.073. Interestingly, both Bio-Bert and ClinicalBert continued to perform well under limited data conditions, demonstrating good generalizability. In 5-shot and zero-shot scenarios, ClinicalBert excelled, achieving impressive F1 and Brier scores of 0.486/0.156 and 0.334/0.181, respectively. This highlights its strength in low-resource settings. Notably, DistilBERT consistently demonstrated the weakest performance across all data, averaging F1 scores of around 0.19 and Brier scores of around 0.19. In both 15-shot and zero-shot conditions, it exhibited a concerning tendency to predict only one class, effectively ignoring the other.

On the MT dataset, similar trends emerge as observed with SOAP. Table \ref{tab:Combined_Tables} showcases our model's evaluation. Both Bio-Bert and ClinicalBert demonstrate strong performance compared to others. Notably, Bio-Bert achieves the highest F1 scores across all data splits: 0.674 (full data), 0.479 (5-shot), and 0.289 (zero-shot). Alongside these, it exhibits competitive Brier scores of 0.096, 0.158, and 0.177, respectively. Interestingly, in the 15-shot scenario, ClinicalBert takes the lead with an F1 score of 0.577 and a Brier score of 0.129, slightly surpassing BioBert. Similar to the SOAP findings, DistilBERT consistently underperforms throughout, mirroring its weakness in low-resource settings. AlBert also shows lower performance across different data splits. As observed before, DistilBERT exhibits its characteristic behaviour of favouring one class, achieving F1 and Brier scores of 0.193 and 0.189 (15-shot) and 0.179 and 0.186 (5-shot).

\begin{figure*}[t!]
    \centering
    \begin{subfigure}[t]{0.32\textwidth}
        \centering
        \includegraphics[height=1.2in]{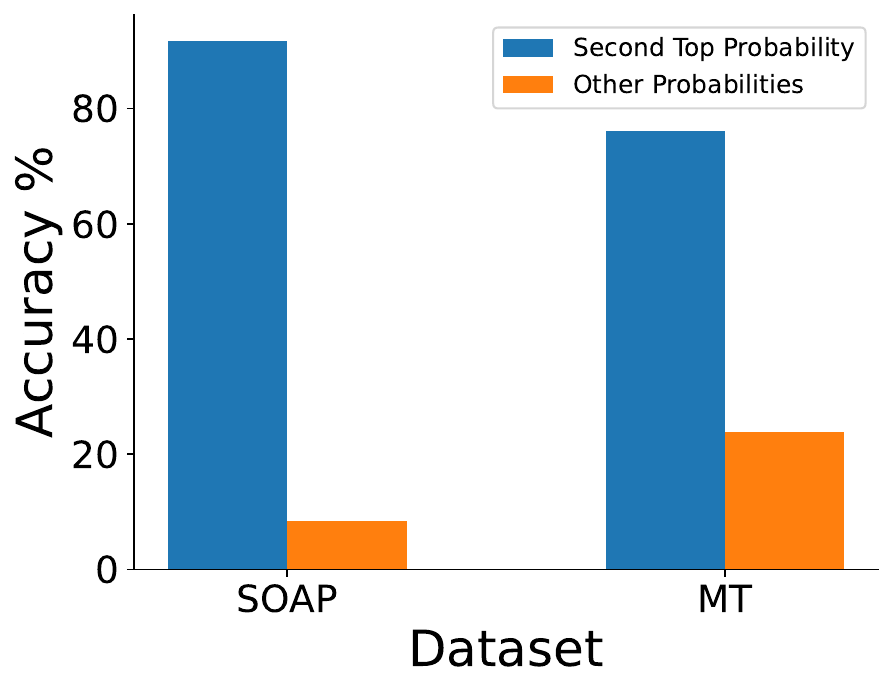}
          \caption{Accuracy of the second-highest probability to other probabilities in incorrect predictions by the Bio-Bert model (15-shot) across multi-class datasets highlighting frequent confusion between the top two classes for incorrect predictions.}
    \label{fig:accuracy_comparison_multi}
    \end{subfigure}%
    ~ 
    \begin{subfigure}[t]{0.32\textwidth}
        \centering
        \includegraphics[height=1.2in]{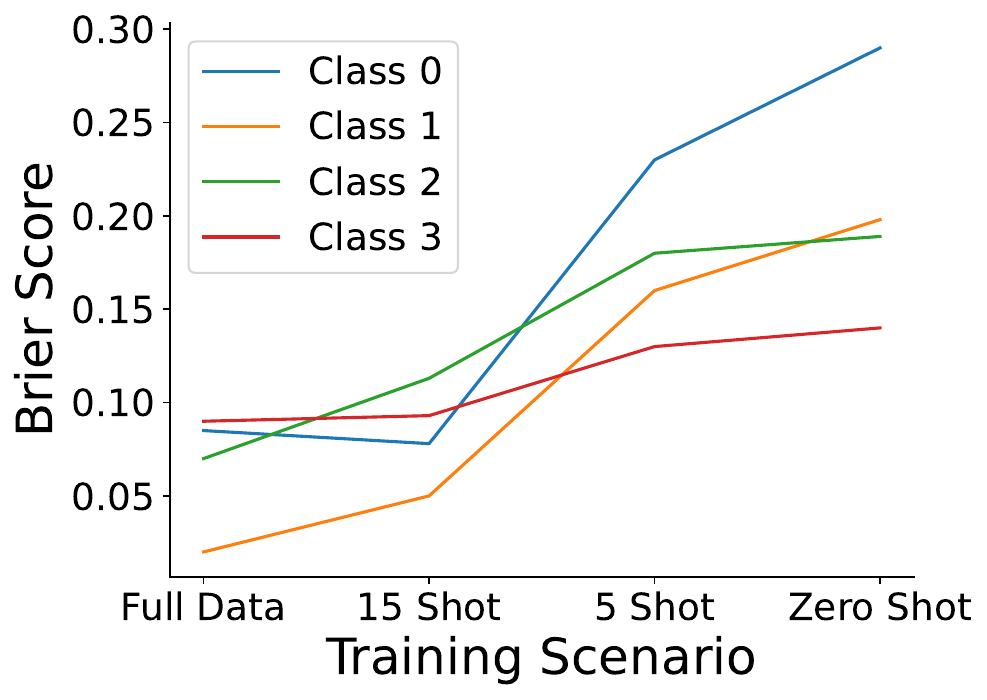}
           \caption{Brier score for each class under different training scenarios for Bio-BERT in the SOAP dataset.}
    \label{fig:brier_score}
    \end{subfigure}
      ~ 
    \begin{subfigure}[t]{0.32\textwidth}
        \centering
        \includegraphics[height=1.2in]{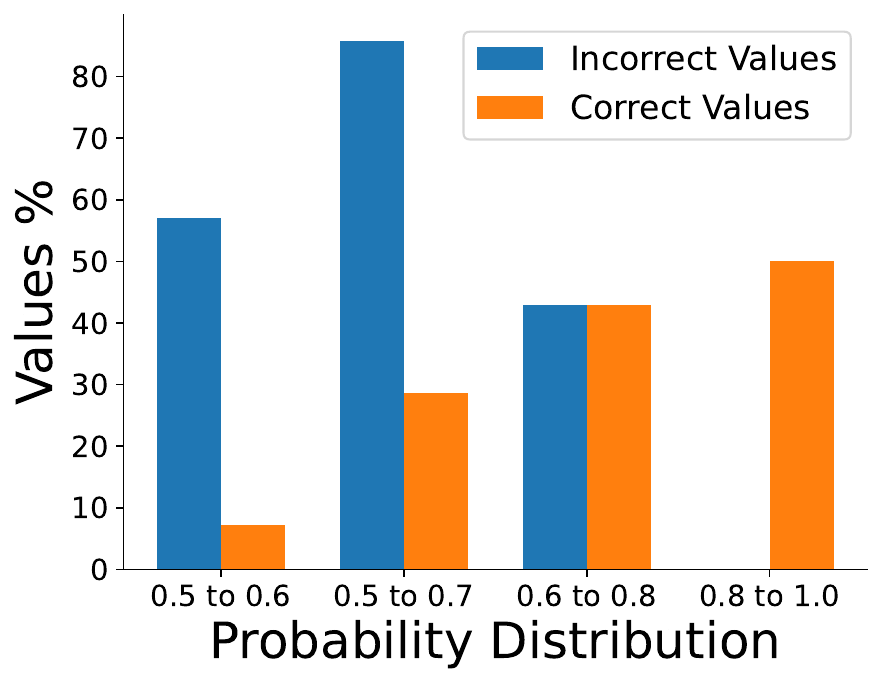}
        \caption{Probability distribution and prediction results for Bio-BERT for 15 Shot ROND Dataset.}
    \label{fig:Prediction Distribution}
    \end{subfigure}
    \caption{Uncertainty analysis across multiple datasets.}
\end{figure*}




The results on the ROND dataset, presented in Table \ref{tab:Combined_Tables}, reveal that ClinicalBert reigns supreme in the full-data scenario. It outperforms all other models in both F1 and Brier scores, achieving impressive marks of 0.916 and 0.056, respectively. However, the picture changes under limited-resource conditions. The results become mixed, with the Bert model generally emerging as the top performer. Specifically, it achieves F1 and Brier scores of 0.788 and 0.114 in the 15-shot scenario, and 0.806 and 0.160 in the 5-shot scenario, demonstrating its relative robustness even with less training data.

\subsection{Overall Discussion}
Our analysis reveals key observations that can aid in enhancing model prediction reliability, particularly in multi-class scenarios where confusion often arises between the top two class probabilities, especially for incorrect predictions. Figure \ref{fig:accuracy_comparison_multi} depicts the trend on multi-class datasets with four classes each. Using the SOAP dataset as an example, analyzing wrong predictions made by Bio-Bert and considering the second-highest probability as correct improves accuracy to 92\%. On the MT dataset, this approach yields a 76\% accuracy. This suggests the model frequently hesitates between the top two classes. By investigating the Brier score (illustrated in Figure \ref{fig:brier_score}), we can further pinpoint the class with the poorest performance. This information can streamline a human-in-the-loop system, allowing it to focus on the specific classes causing confusion rather than reviewing all predictions.

The model's Bayesian architecture shines in its ability to capture uncertainty within its predictions. We demonstrate this through error analysis and confidence estimate modelling. One crucial benefit is facilitating targeted error analysis. By identifying instances where the model is confused and struggles with decision-making, we can prioritize these predictions for further scrutiny. Figure \ref{fig:Prediction Distribution} presents the probability distribution of accurate and inaccurate predictions for the Bio-BERT model in a binary classification setting. The distribution is divided into four segments. As expected, predictions approaching 1 exhibit high confidence, while those near 0.5 reveal greater uncertainty. For a more nuanced view, we zoomed in on the 0.6--0.8 range. The figure shows that highly confident predictions (0.8-1.0) are often correct, while uncertain predictions (0.5-0.7) frequently lead to errors. This underscores the model's ability to flag confusing cases. By leveraging our model, we can efficiently pinpoint these flagged predictions, directing our attention to instances requiring closer examination.

\section{Conclusion}
This paper presented a novel Bayesian deep learning model with kernel dropout, specifically designed to enhance the reliability of predictions in medical text classification tasks. Our model demonstrated significant improvements in both effectiveness and calibration, especially in low-resource settings. While further research is needed to address limitations such as performance in other domains and explore different applications, this work highlights the potential of Bayesian deep learning models with uncertainty quantification to build trust and improve outcomes in AI-driven healthcare. By combining the power of deep learning with the rigour of Bayesian inference, we can pave the way for more reliable and interpretable AI tools that assist medical professionals in delivering better patient care.

\bibliographystyle{splncs04}
\bibliography{sbibl}

%





\end{document}